\documentclass[final,5p,times,twocolumn]{elsarticle}

\usepackage{lineno}
\modulolinenumbers[5]

\usepackage[dvipsnames]{xcolor}
\usepackage{multirow}

\newcommand*\rotHalf{\rotatebox{0}}

\usepackage{amssymb}
\usepackage{latexsym}
\usepackage{algpseudocode}
\usepackage{blindtext}
\usepackage{algorithm}
\usepackage{times}
\usepackage{epsfig}
\usepackage{graphicx}
\usepackage{amsmath}
\usepackage{textcase}
\usepackage{comment}
\usepackage{breqn}
\usepackage{ragged2e}
\usepackage[normalem]{ulem}
\usepackage{caption, booktabs}

\usepackage{upgreek}
\usepackage{array}
\usepackage{textcomp}
\newcommand*\rot{\rotatebox{90}}
\newcolumntype{P}[1]{>{\centering\arraybackslash}p{#1}}
\useunder{\uline}{\ul}{}
\usepackage{balance}
\usepackage{lastpage}
\usepackage[pagebackref=true,breaklinks=true,letterpaper=true,colorlinks,bookmarks=false]{hyperref}

\usepackage[pagebackref=true,breaklinks=true,letterpaper=true,colorlinks,bookmarks=false]{hyperref}

\usepackage{todonotes}
\usepackage{xparse}
\usepackage{array}
\newcolumntype{L}[1]{>{\raggedright\let\newline\\\arraybackslash\hspace{0pt}}m{#1}}
\newcolumntype{C}[1]{>{\centering\let\newline\\\arraybackslash\hspace{0pt}}m{#1}}
\newcolumntype{R}[1]{>{\raggedleft\let\newline\\\arraybackslash\hspace{0pt}}m{#1}}
\newcolumntype{J}[1]{>{\justifying\let\newline\\\arraybackslash\hspace{0pt}}m{#1}}

\usepackage{enumitem}
\usepackage{float}

\newcommand{\myComment}[1]{}

\begin{document}
\begin{frontmatter}

\title{Distribution Regularized Self-Supervised Learning for Domain Adaptation of Semantic Segmentation}

\author[1]{Javed~Iqbal\corref{cor}}
\cortext[cor]{Corresponding author}
\ead{javed.iqbal@itu.edu.pk}

\author[1]{Hamza~Rawal}
\ead{mscs18004@itu.edu.pk}

\author[1]{Rehan~Hafiz}
\ead{rehan.hafiz@itu.edu.pk}

\author[2]{Yu-Tseh Chi}
\ead{jchi@fb.com}

\author[1]{Mohsen~Ali}
\ead{mohsen.ali@itu.edu.pk}

\address[1]{Information Technology University, Lahore, 54000, Pakistan}
\address[2]{Facebook, 1 Hacker Way, Menlo Park, CA 94025, USA.}



\begin{abstract}

This paper proposes a novel pixel-level distribution regularization scheme (DRSL) for self-supervised  domain adaptation of semantic segmentation.
In a typical setting, the classification loss forces the semantic segmentation model to greedily learn the representations that capture inter-class variations in order to determine the decision (class) boundary. 
Due to the domain-shift, this decision boundary is unaligned in the target domain, resulting in noisy pseudo labels adversely affecting self-supervised domain adaptation.
To overcome this limitation, along with capturing inter-class variation, we capture pixel-level intra-class variations through class-aware multi-modal distribution learning (MMDL). Thus, the information necessary for capturing the intra-class variations is explicitly disentangled from the information necessary for inter-class discrimination.  
Features captured thus are much more informative, resulting in pseudo-labels with low noise. 
This disentanglement allows us to perform separate alignments in discriminative space and multi-modal distribution space, using cross-entropy based self-learning for former. 
For later, we propose novel stochastic mode alignment method, by explicitly decreasing the distance between the target and source pixels that map to the same mode.
 The distance metric learning loss, computed over pseudo-labels and backpropagated from multi-modal modeling head, acts as the regularizer over the base network shared with the segmentation head.
 The results from comprehensive experiments on synthetic to real domain adaptation setups, i.e.,  GTA-V/SYNTHIA to Cityscapes, show that DRSL outperforms many existing approaches  (a minimum margin of 2.3\% and 2.5\% in mIoU for SYNTHIA to Cityscapes).

\end{abstract}


\end{frontmatter}

\begin{figure*}[t]
 	\centering
 	\includegraphics[width= \textwidth]{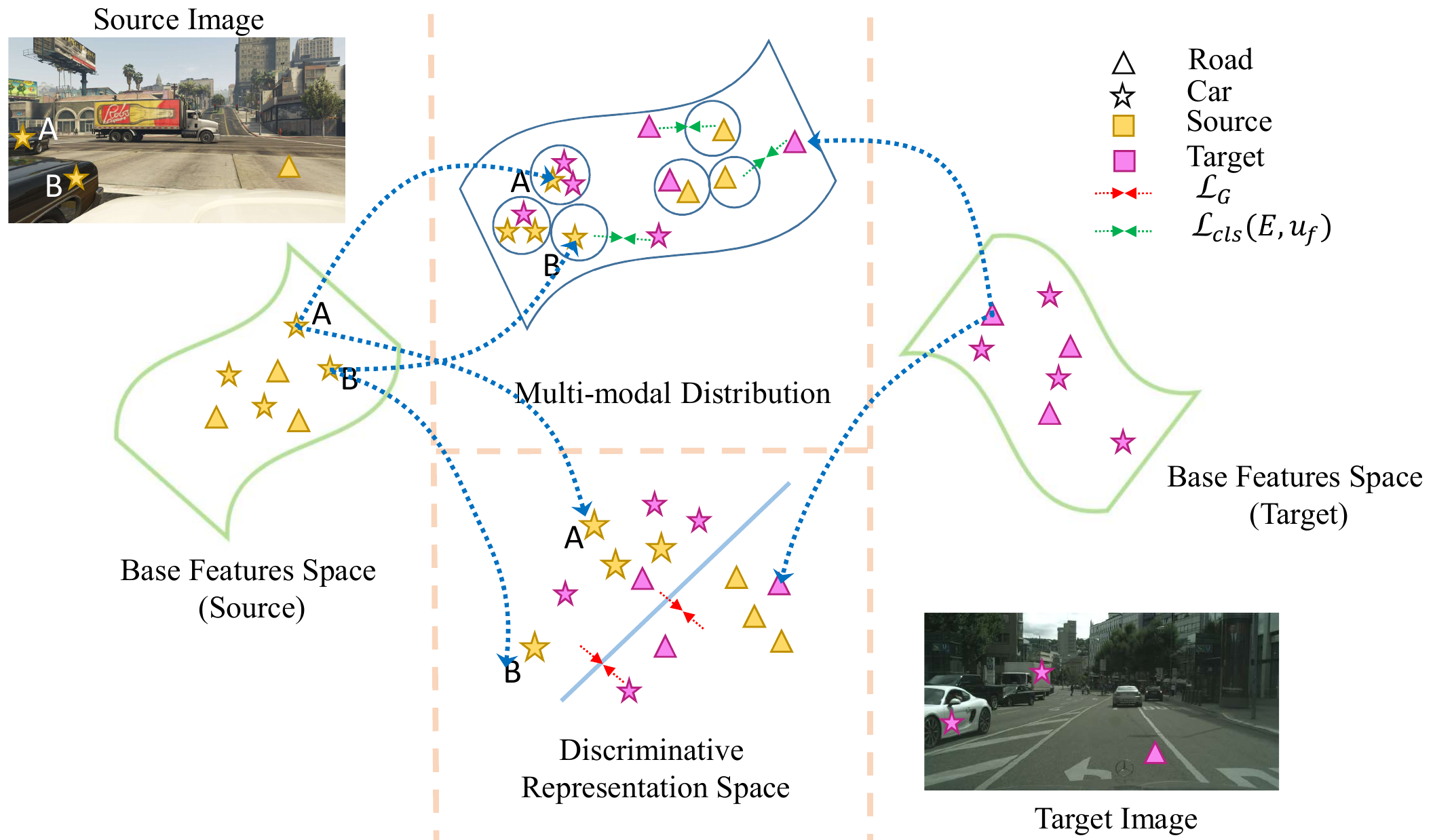}
 	\scriptsize
 	\caption{Separately capturing pixel-level intra-class variations and inter-class discriminative information, enables us to perform different alignment operations, i.e. \textit{class aware mode alignment} $\&$ \textit{cross-entropy based decision boundary alignment}. A $\&$ B are consistent in-term of being on same side of decision boundary, but variant enough to map to different modes.
 	} 
 	\label{img:teaser}
\end{figure*}

\section{Introduction}

In recent years, deep neural network based semantic segmentation models have achieved considerable success. This success is much reliant on the large pixel-level annotated dataset over which these models are trained. 
However, like many other deep neural network based models, semantic segmentation models suffer from considerable performance degradation when tested on images from the domain different than then one used in training. 
This problem, attributed to the domain shift, is exacerbated in semantic segmentation algorithms since many of them are trained on the synthetic dataset, due to lack of large real-world annotated datasets, and are tested over the real-world images. 
Retraining or fine-tuning for new domains is expensive, time consuming, and in many cases not possible due to the large number of ever-changing domains, especially in case of autonomous vehicles, and unavailability of annotated data.

To overcome domain shift, unsupervised domain adaptation (UDA), has been employed with reasonable success \cite{zou2018unsupervised,zou2019crst, mlsl2020}, but state-of-the-art is still lacking desired accuracy. 
Many unsupervised domain adaptation algorithms for semantic segmentation \cite{hoffman2017cycada, chen2017no,clan_2019_CVPR, Yunsh2019bidirect, structure_2019_CVPR, dada_2019_ICCV, iqbal2022leveraging} perform global marginal distribution alignment through adversarial learning to translate the input image or feature volume or output probability tensor from one domain to other. 
The adversarial loss looks at the whole tensor (image/feature or output probability) even when the objective is to improve the pixel-level label assignments \cite{clan_2019_CVPR}, more-over aligning marginal distributions does not guarantee preserving the discriminative information across the domain \cite{zhang2019category}.
Self-supervised learning methods \cite{mlsl2020, pan2020unsupervised, Lian_2019_pycda, LSE_2020_Naseer, zou2019crst, Yunsh2019bidirect,munir2021ssal} (either independently or along with adversarial learning ) try to overcome this challenge by back-propagating the cross-entropy loss computed over pixel-level pseudo-labels generated by the source model. 
Quality of these pseudo-labels is dependent upon the generalization capacity of the classifier and effects overall adaptation process.
The deep neural network based semantic segmentation model when trained by minimizing cross-entropy loss, greedily learns representations that capture inter-class variations. 
When optimally trained these inter-class variations should help map accurate decision boundary, projecting pixels from different classes to different sides of it (decision boundary).
However, due to the domain shift, the decision boundary is not aligned in target domain, resulting in noisy pseudo-labels leading to poor self-supervised domain adaptation. 

Previous works \cite{chen2020homm, kumagai2019unsupervised} have shown discriminative clustering on target data and moment matching across domains helps in adaptation .
CAG-UDA \cite{zhang2019category}  $\&$ \cite{Deng_2019_ICCV} tried to align the class aware cluster centers across domains for better adaptation. 
However, visual semantic classes exhibit large set of variations, due to difference in texture, style, color, pose, illumination etc.. These variations are generally assumed to be across instance, e.g. two different types of cars, but they do manifest frequently in the same instance too, e.g. pixels belonging to different road locations or to different parts of car. 
Class aware single cluster based alignment might align centers of the source and target domain without aligning overall distribution, leaving classes with large variations vulnerable to misclassification in target domain. 
Learning to capture intra-class variations by representing each class with multiple modes and aligning the modes across domain might overcome these challenges. 

Therefore, we propose a novel class aware multi-modal distribution alignment method for unsupervised domain adaption of semantic segmentation model.
We combine together the ideas of distribution alignment and pseudo-label based adaptation, however, instead of just using discriminatively learned features during the adaptation, we explicitly learn representations separately.
In addition to learning the inter-class variation through minimizing cross-entropy loss, i-e the pixel-level intra-class features variations are captured by learning a multi-modal for each class (Fig. \ref{img:teaser}), resulting in a much more generalized representation. 
Both of these tasks have competing requirements, minimizing cross entropy loss results in learning inter-class discriminative representation along with intra-class consistency. Whereas multi-modal distribution learning intends to preserve information that can model intra-class variations. 
We disentangle these two information requirements by developing class-aware multi-modal distribution learning (MMDL) module , parallel to standard segmentation head. 
MMDL extracts the spatially low-resolution  feature volume from the encoding block and maps to the spatially high-resolution embedding. 
Class aware multi-modal modeling is performed over these embedding using Distance metric learning \cite{repmet2019}.
Since both of these heads share the backbone, simultaneously decreasing loss on both act as a regularizer over the learned features, resulting in the less noisy pseudo-labels.
During domain adaptation, the high quality pseudo-labels allow us to learn domain-invariant class discriminative feature representations in the discriminative space. 
At the same time,  stochastic mode-alignment is performed across domains, by minimizing distance between representation of source pixels and target pixels mapping to same mode; thus preserving intra-class variations. 
Modes themselves are updated by increasing the posterior probability of target pixel belonging to the mode identified closest to target. 
During adaptation too, these losses computed paralelly act as regularizer over each other, hence dampening each others noise.

Our contributions are summarized as follows. 
First, we propose a multi-modal distribution alignment strategy for the self-supervised domain adaptation. 
By designing a multi-modal distribution learning (MMDL) module parallel to standard segmentation head, with shared backbone, we disentangle inter-class discriminative and intra-class variation information; allowing them to be used during adaptation separately. 
We show that due to regularization of MMDL, the pseudo-labels generated over target domain are more accurate. Lastly, to perform stochastic mode alignment, we  introduce the \textit{cross domain consistency loss}. 
We present state-of-the-art performance for benchmark synthetic to real, e.g., GTA-V/SYNTHIA to Cityscapes adaptation.

\begin{figure*}[t]
 	\centering
 	\includegraphics[width=1 \textwidth]{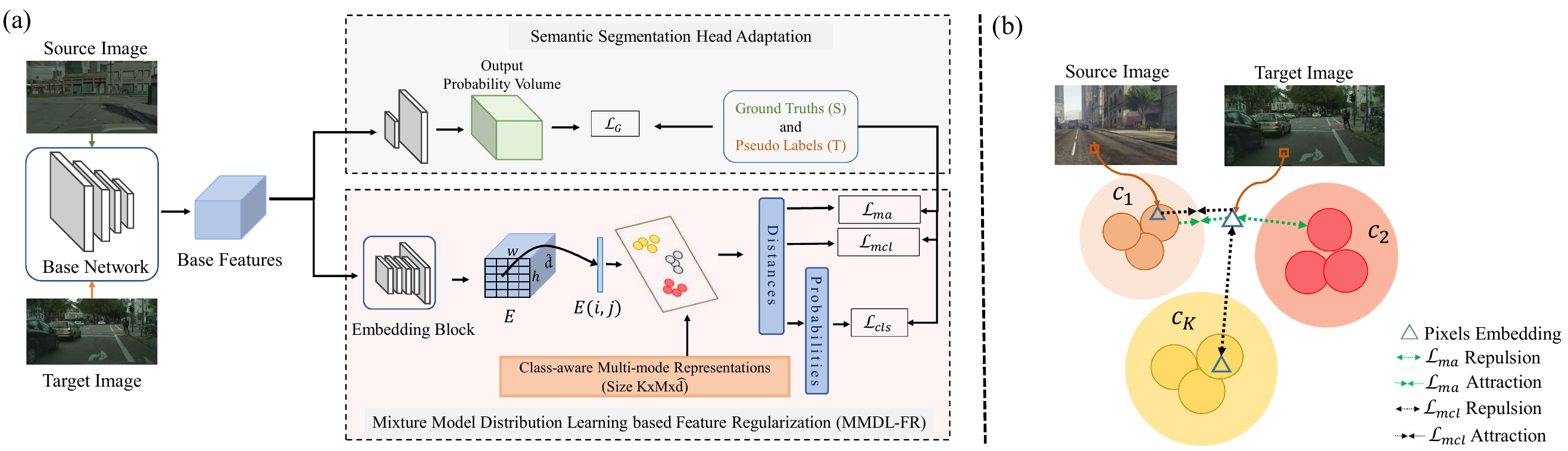}
 	\scriptsize
 
 	\caption{The proposed DRSL approach (a) Base features extracted from the base network are used for two separate tasks. The MMDL-FR module captures intra-class variations through multi-modal distribution learning. Semantic Segmentation head estimates the discriminative class boundaries necessary for the primary segmentation task. This disentanglement allows us simultaneous alignment in discriminative and multi-modal space, allowing MMDL-FR module to act as a regularizer over the Segmentation Head. (b) The proposed \textit{Stochastic mode alignment}: Minimizing $\mathcal{L}_{mcl}$ brings the source and target embeddings of the same mode of the same class closer than any source pixel’s embedding belonging to different class. $\mathcal{L}_{ma}$ decreases the in-mode variance for the target samples by forcing them to come closer to the assigned mode and move away from other class's  modes. 
 	}
 	\label{img:model}
\end{figure*}

\section{Related Work}
The domain shift between testing and training data deteriorates the model performance in most of the computer vision tasks like classification \cite{tzeng2017adversarial, pinheiro2018unsupervised, xu2019larger, deng2019cluster, belal2021knowledge, schrom2021improved}, object detection \cite{chen2018domain, khodabandeh2019robust,hsu2020progressive} and semantic segmentation \cite{chen2017no, chen2017road, clan_2019_CVPR, curr2017_ICCV, dai2019curriculum, hoffman2017cycada, iqbal2020weakly, LSE_2020_Naseer, yang2021exploring}. In this work, we focus on the domain shift problem for semantic segmentation with self-supervised learning. Our work is related to semantic segmentation, domain adaptation, and self-supervised learning.

\noindent \textbf{Domain Adaptation for Semantic segmentation:}

Recent works \cite{vu2019advent, mlsl2020, LSE_2020_Naseer, chen2017road, chen2017no, tsai2018learning, Lian_2019_pycda, iqbal2020weakly, zou2018unsupervised, Cordts2016Cityscapes, dada_2019_ICCV, guo2021metacorrection} aiming to minimize the distribution gap between source and target domains are focused in two main directions.  1) adversarial learning and, 2) self-supervised learning for unsupervised domain adaptation (UDA) of semantic segmentation. 

\noindent \textbf{Adversarial Domain Adaptation:} Adversarial learning is the most explored area for output space \cite{tsai2018learning, wang2020differential, vu2019advent, pan2020unsupervised,dada_2019_ICCV}, latent/feature space \cite{chen2017no, mancini2018boosting} and input space adaptation \cite{hoffman2017cycada, clan_2019_CVPR, zhang2018fully, Yunsh2019bidirect}. We briefly describe the feature space/feature alignment, as our work is related to it. 
The authors in \cite{kim2020learning, hoffman2017cycada, zhang2020towards} used adversarial loss to minimize the distribution gap between the high level features representations of the source and target domain images. 
However, these methods do not align class-wise distribution shifts but instead match the global marginal distributions. To overcome this, \cite{chen2017no, clan_2019_CVPR} combined category level adversarial loss (by defining class discriminators) with domain discriminator at feature space. \cite{iqbal2020weakly} tried to regularize the segmentation network using weak labels along with latent space marginal distribution alignment for domain adaptation of semantic segmentation. 
\textcolor{black}{Similarly, the authors in \cite{yang2021exploring} investigated the robustness of the UDA of semantic segmentation and proposed a self-training augmented adversarial learning to improve the robustness to adversarial examples. Their approach resulted better performance in the presence of adversarial examples, however, reducing the performance over normal input images. 
}

\noindent \textbf{Self-supervised learning:}
Self-supervised learning for UDA is recently studied for major computer vision tasks like semantic segmentation and object detection \cite{tri2018fully, mlsl2020, khodabandeh2019robust, Lian_2019_pycda}. 
The authors in \cite{zou2018unsupervised} proposed a self-paced self-training approach by generating class balanced pseudo-labels and class spatial priors extracted from the source dataset used to condition the pseudo-label generation. Zou et al. \cite{zou2019crst} extended the \cite{zou2018unsupervised} with confidence regularization strategies and soft pseudo-labels for self-training based UDA for semantic segmentation. LSE \cite{LSE_2020_Naseer} further worked with self-generated scale-invariant examples and entropy based dynamic selection for self-supervised learning. 
\textcolor{black}{The authors in \cite{guo2021metacorrection} proposed a domain-aware meta-learning approach (MetaCorrection) to correct the segmentation loss and condition the pseudo-labels based on noise transition matrix. They report considerable mIoU gain especially when applied on pre-adapted model. 
}
In this work, we exploit a strategy similar to \cite{zou2018unsupervised} to generate pseudo-labels for target domain images during adaptation.

\noindent \textbf{Clustering Based Features Regularization:} 
Some previous works also explored the effect of discriminative clustering on target data and moment matching across domains for target data adaptation \cite{chen2020homm, kumagai2019unsupervised}.
Recently, \cite{zhang2019category, Deng_2019_ICCV} tried to define category anchors on the last feature volume of the segmentation model to align class aware centers across the source and target domains. Tsai et al. \cite{tsai2019domain} tried to match the clustering distribution of discriminative patches from source and target domain images. Similarly, \cite{mlsl2020} and \cite{Lian_2019_pycda} exploited latent space and output space respectively by defining category based classification modules, forcing towards class-aware adaptation. 
However, these methods do not explore the intra-class variations present in source or target data but instead
leverage the discriminative property to align the inter-class clusters. We specifically focus to capture the intra-class variations present in the source and target data by learning class-aware mixture models to help the adaptation.

\section{Distribution Regularised Self-supervised Learning}
\label{sec:method}
In this section, we provide details of our distribution regularized self-supervised learning (DRSL) architecture. It employs DeepLab-v2 \cite{chen2018deeplab} as a baseline and embeds new components that enable the semantic segmentation model to be robust to domain shift.


\subsection{Preliminaries}

For supervised semantic segmentation, we have access to source domain images $\{\mathrm{x_s, y_s}\}$ from $X_s \in \mathbb{R} ^{H\times W\times 3}$ with corresponding ground truth labels $Y_s \in \mathbb{R} ^{H\times W\times K}$. The $\{H, W\}$ shows the width and height of source domain images and $K$ shows the number of classes. Let $\mathcal{G}$ be a segmentation model with weights $\mathrm{w_g}$ that predicts the $K$ channel softmax probability outputs. For a given source image $\mathrm{x_s}$, the segmentation probability vector of class $c$ at any pixel location (${i,j}$) is obtained as $p(c|\mathrm{x_s, w_g})_{i,j} = \mathcal{G}(\mathrm{x_s})_{i,j}$
%
%
For fully labeled source data, the network parameters $\mathrm{w_g}$ are learned by minimizing the cross entropy loss (Eq. \ref{eqn:2}),
\begin{equation}
\small
    \mathcal{L}^s_{seg} (\mathrm{x_s, y_s}) = -\sum_{i=1}^H \sum_{j=1}^W \sum_{c=1}^K \mathrm{y_s^{(c,i,j)}} ~\log(p(c|\mathrm{x_s, w_g})_{c,i,j})
\label{eqn:2}
\end{equation}
where $\mathcal{L}^s_{seg}$ is the source domain segmentation loss. 
For unsupervised domain adaptation of the target domain, we have access to the target domain images $\{\mathrm{x_t, -}\}$ from $X_t \in \mathbb{R} ^{H_t\times W_t\times 3}$ with no ground truths available.
Thus, we adapt the iterative process used by \cite{mlsl2020, zou2018unsupervised} to first generate pseudo-labels $\mathrm{\hat{y}_t}$ using the source trained model and then fine-tune the source trained model on target data using Eq. \ref{eqn:3}.  
\begin{equation}
\small
    \mathcal{L}^t_{seg} (\mathrm{x_t, \hat{y}_t}) = -\sum_{i=1}^{H_t} \sum_{j=1}^{W_t} \mathrm{b^{(i,j)}_t} \sum_{c=1}^K \mathrm{\hat{y}_t^{(c,i,j)}} \log(p(c|\mathrm{x_t, w_g})_{c,i,j})
\label{eqn:3}
\end{equation}
where $\mathcal{L}^t_{seg}$ is the segmentation loss for target domain images 
with respect to generated pseudo-labels $\mathrm{\hat{y}_t}$. $\mathrm{b_t}$ represents a binary mask with same resolution as $\mathrm{\hat{y}_t}$ to back-propagate loss for pixels which are assigned pseudo-labels.  
The total loss for the segmentation model is the combination of true labels based source domain loss and pseudo-labels based target domain loss and is given by Eq. \ref{eqn:loss_seg_total},
\begin{equation}
\small
    \mathcal{L}_{\mathcal{G}}(\mathrm{x_s, y_s,x_t, \hat{y}_t}) = \mathcal{L}^s_{seg} (\mathrm{x_s, y_s}) + \mathcal{L}^t_{seg} (\mathrm{x_t, \hat{y}_t}) 
\label{eqn:loss_seg_total}
\end{equation}
%

\subsection{Multi-Modal Distribution Learning}
\label{sec:drsl}
We propose to learn the complex intra-class variations through a multi-modal distribution learning (MMDL) framework where instead of a single cluster/anchor, each class is represented by multiple modes. This diverse representation of each class is used in the adaptation process to align the domains on fine-grained level. Furthermore, we disentangle the task of learning these intra-class variations (MMDL) from the main segmentation task by designing a separate module for it called multi-modal distribution learning based feature regularization (MMDL-FR). The proposed MMDL-FR module is model agnostic and can be appended at the encoder of any segmentation network.

The MMDL-FR module consists of mixture models based per-pixel classification augmented with distance metric learning (DML) based per-pixel embedding block.
The input of the MMDL-FR module is the feature volume $F \in \mathbb{R} ^{h\times w\times d}$, where $\{h, w, \text{and} ~d\}$ shows the spatial height, width and depth of the encoder output (base features) as shown in Fig. \ref{img:model}(a). 
The embedding block is comprised of 4 fully convolutional layers with different dilation rates (similar to ones used in the last layer of the segmentation network) followed by an upsampling layer. 
The output of the embedding block $\mathcal{E}$ is a feature volume $E = \mathcal{E}(F) \in \mathbb{R} ^{h_o\times w_o\times \hat{d}}$, where $(h_o, w_o)=(H/2, W/2)$ (Sec.\ref{sec:ablation}) and $d>>\hat{d}$ for any randomly selected source image.

To train the MMDL-FR module, we adapt a formulation similar to \cite{repmet2019}.
For each class $c$, a multi-modal distribution with $M$ number of modes is learned. 
Let $e=E(i,j)$ be embedding for location $(i, j)$, a vector $V^{c}_m$ represent the center of the mode $m, (m=1,..,M)$ of the class $c, (c=1,...,K)$ of the mixture models. In this work, these mode centers are formulated as the weights of a fully connected layer with size $K \cdot ~M \cdot ~\hat{d}$, and are reshaped into $(K \times M) \times \hat{d}$ producing $K \times M$ matrix for each input embedding vector $e$. This simple method makes it easy to flow back gradients to the fully connected layer and learn the segmentation backbone during training. 
To compute the classification probability for each embedding vector $e$, we compute the euclidean distance $D^{c}_m(e) = ||e-V^{c}_m||^2_2$ between $e$ and representative $V^{c}_m$ and compute the posterior probabilities $q^c_m(e) \propto exp(- (D^{c}_m(e))^2 / 2\sigma^2)$,
%
%
%
\textcolor{black}{where $\sigma^2$ is the variance of each mode and is set to 0.5.}
For class $c$ posterior probability, we take the maximum over $M$ modes of class $c$ as, $Q(C=c|e)=\max_{m=1,...,M} q^c_m(e)$, 
%
%
where C = c shows class c. 

\noindent \textbf{Loss Functions: }
To train the MMDL-FR module, two losses are used, i.e., triplet loss and the cross entropy loss. The triplet loss for \textit{embedding block} is defined by Eq. \ref{eqn:6}, 
\begin{equation}
\small
    \mathcal{L}_{emb}(E) =\sum_{e \in E} |\min_m D^{c^*}_m(e) - \min_{m, c^* \neq c}D^{c}_m(e) + \alpha|_+
\label{eqn:6}
\end{equation}
where $|.|_+$ is the Relu function and $\alpha$ is the minimum margin between the distance of an embedding $e$ to the closest mode representative $V^{c^*}_m$ of the true class $c^*$,  and distance of embedding $e$ to the closest mode representative of the incorrect class $V^{c}_m$.
Similarly, the cross entropy loss for mixture models based classification is given by Eq. \ref{eqn:7},
\begin{equation}
\small
    \mathcal{L}_{cls} (\mathrm{E, u_f}) = - \sum_{e \in E} \sum_{c=1}^K \mathrm{u_f^{(c)}} \log(Q(C=c|e))
\label{eqn:7}
\end{equation}
where $\mathrm{u_f^{(c)}}$ is the embedding classification label obtained from $\mathrm{y_s^{c}}$ or $\mathrm{\hat{y}_t^{c}}$ for class $c$.  
The triplet loss enforces the embedding block to learn representation that capture intra-class variation information, while cross entropy loss pushes them to not lose necessary class-specific information. 
Due to these two losses, the MMDL-FR module acts as a regularizer at latent space over the shared backbone, so that the shared features are much more informative if only segmentation head is used. 

\subsection{Stochastic Mode Alignment}
One of the characteristic of domain generalization will be that the multi-modal distribution learning over one domain should result in the modes which are very close to the modes learned in the other domain. 
However, due to the domain shift, this is not generally true. 
That is in the target domain, the features of pixels assigned pseudo-label $c$ might not be closer to the any of the modes belonging to the class $c$.
In addition, features in target domain mapping to same mode might not be closer to each other, resulting in low posterior probability. 
We minimize two loss functions to perform \textit{stochastic mode alignment}. 

For first, we apply \textit{domain invariant consistency loss}, ensuring that features of pixels mapped to same modes of same class should be near to each other regardless of the domain the are sampled from. 
Assume a batch consisting of arbitrary number of source and target images, $\{(x_s^i, y_s^i)|i=0,1\dots,N_s,(x_t^i,\hat{y}_t^i)|i=0,1\dots,N_t \}$, where $\hat{y}_t^i$ are the pseudo-labels assigned to $x_t^i$. 
Embedding $E_t^i=\mathcal{E}(x_t^i)$  and $E_s^i=\mathcal{E}(x_s^i)$ are computed for all the target and source images in the batch. 
We randomly sample $N_e$ number of embedding, $\{e_t^i|i=0,1\dots,N_e\}$ from $\{E_t^i|i=0,1\dots,N_t \}$, choosing only from the ones having valid pseudo-label.

For \textit{domain invariant consistency}, we create a triplet $(e_t^i, e_s^i, \hat{e}_{s}^i)$ such that pseudo-label of $e_t^i$ and ground-truth label of $e_s^i$ is same class $c$, and both map to same mode $m$ of class $c$. $\hat{e}_{s}^i$ on the other hand is source pixel's embedding of any class $c^{+} \ne c$.
This loss when minimized brings $e_t^i$ closer to $e_s^i$ than any source pixel's embedding belonging to different class. 
\begin{equation}
\small
    \mathcal{L}_{mcl} = \sum_{i}^{N_e} | || e_t^i - e_s^i ||_2^2 - || e_t^i - \hat{e}_{s}^i ||_2^2 + \alpha1|_+
\label{eqn:ada}
\end{equation}
Note: we could have chosen most closest source sample as negative, however, this would have been computationally prohibitive. Margin, $\alpha1$, is set to 1, for all experiments. 

The in-mode variance for the target samples is decreased by forcing them to come closer to the assigned mode and move away from the modes of the other classes. We sample $T_e$ embeddings per image per class from both source and target images and create set $E_s$ and $E_t$ respectively. Eq. \ref{eqn:6-sma} minimizes the triplet loss for both the source and target embeddings simultaneously. 
\begin{equation}
\small
    \mathcal{L}_{ma}(E_s, E_t) = \frac{1}{T^s_e} \mathcal{L}_{emb}(E_s) + \frac{1}{T^t_e} \mathcal{L}_{emb}(E_t)
\label{eqn:6-sma}
\end{equation}
where $T^s_e$ and $T^t_e$ represent cardinality of $E_s$ and $E_t$, which might be different since samples from all classes might not be available.

\subsection{Total Loss for Training and Adaptation}
The DRSL model is trained using the combination of segmentation losses, mode consistency loss and MMDL-FR module losses. 
Let $\mathcal{L}_{cls}^s$ and $\mathcal{L}_{cls}^t$ represent call to Eq. \ref{eqn:7} using sourse and target embeddings respectively.   
The source model with MMDL module is trained using Eq.\ref{eqn:loss-src}. 
\begin{equation}
\begin{split}
\small
   \mathcal{L}_{src} = \mathcal{L}^s_{seg} + \beta ~\mathcal{L}_{emb} +\eta \mathcal{L}_{cls}^s
\label{eqn:loss-src}
\end{split}
\end{equation}
During adaptation to target domain the loss functions in Eq.\ref{eqn:drsl} and Eq.\ref{eqn:drsl+} are used.
\begin{equation}
\begin{split}
\small
   \mathcal{L}_{DRSL} = \mathcal{L}_{\mathcal{G}} + \beta ~\mathcal{L}_{ma} +\eta (\mathcal{L}_{cls}^s+\mathcal{L}_{cls}^t)
\label{eqn:drsl}
\end{split}
\end{equation}
\begin{equation}
\begin{split}
\small
   \mathcal{L}_{DRSL+} = \mathcal{L}_{\mathcal{G}} + \beta ~\mathcal{L}_{ma} +\eta (\mathcal{L}_{cls}^s+\mathcal{L}_{cls}^t) + \gamma ~\mathcal{L}_{mcl}
\label{eqn:drsl+}
\end{split}
\end{equation}
where, $\beta$, $\eta$ and $\gamma$ are hyper-parameters to limit the effect of MMDL-FR module loss values. 


\begin{table*}[h]
\centering
\caption{Semantic segmentation performance for GTA-V to Cityscapes adaptation. The abbreviations “$A_I$”, “$A_F$” and “$A_O$” stand for adversarial training at input space, latent space, and output space. Similarly, “$S_T$” represents self-supervised learning.}
\resizebox{\textwidth}{!}{
\begin{tabular}{l|c|c|ccccccccccccccccccc|c|c}
\hline 
\multicolumn{23}{c}{GTA-V $\rightarrow$ Cityscapes}\\
\hline
Methods & \rot{Baseline}    & \rot{Appr.} & \rot{Road}  & \rot{Sidewalk} & \rot{Building} & \rot{Wall}  & \rot{Fence} & \rot{Pole}  & \rot{T. Light} & \rot{T. Sign} & \rot{Veg.} & \rot{Terrain} & \rot{Sky}   & \rot{Person} & \rot{Rider} & \rot{Car}   & \rot{Truck} & \rot{Bus}   & \rot{Train} & \rot{M.cycle} & \rot{Bicycle} & \rot{mIoU}& \rot{mIoU Gain}  \\ \hline \hline

Source \cite{chen2018deeplab} & \multirow{8}{*}{\rotHalf{DeepLab-v2}}        & -& 75.8    & 16.8    & 77.2    & 12.5    & 21.0    & 25.5    & 30.1    & 20.1    & 81.3    & 24.6    & 70.3    & 53.8    & 26.4    & 49.9    & 17.2    & 25.9    & 6.5    & 25.3    & 36.0    & 36.6 & - \\

MinEnt \cite{vu2019advent}         &  & $A_O + S_T$         & 86.6    & 25.6    & 80.8    & 28.9    & 25.3    & 26.5    & 33.7    & 25.5    & 83.3    & 30.9    & 76.8    & 56.8    & 27.9    & 84.3    & \textbf{33.6}    & 41.1    & 1.2    & 23.9    & 36.4    & 43.6    &7.0\\

FCAN \cite{zhang2018fcan}         &  & $A_I + A_O$         & -    & -    & -    & -    & -    & -    & -    & -    & -    & -    & -    & -    & -    & -    & -    & -    & -    & -    & -    & 46.6    &10.0\\

IntraDA \cite{pan2020unsupervised}        &  & $A_O + S_T$      & 90.6     &  37.1     &  82.6     &  30.1     &  19.1     &  29.5     &  32.4     &  20.6     &  {\ul 85.7}     &  {\ul 40.5}     &  79.7     &  58.7     &  {\ul 31.1}     &  \textbf{86.3}     &  31.5     &  {\ul 48.3}     &  0.0     &  30.2     &  35.8     &  46.3     & 9.7 \\
PyCDA \cite{Lian_2019_pycda}        &  & $S_T$      & 90.5  & 36.3  & \textbf{84.4}  & {\ul 32.4}  & \textbf{28.7}  & 34.6  & 36.4  & 31.5  & \textbf{86.8}  & 37.9  & 78.5  & 62.3  & 21.5  & {\ul 85.6}  & 27.9  & 34.8  & {\ul 18.0}  & 22.9  & \textbf{49.3}  & 47.4 & 10.8 \\

LSE \cite{LSE_2020_Naseer}        &  & $S_T$      & 90.2    & 40.0    & {\ul 83.5}    & 31.9    & {\ul 26.4}    & 32.6    & 38.7    & 37.5    & 81.0    & 34.2    & \textbf{84.6}    & 61.6    & \textbf{33.4}    & 82.5    & {\ul32.8}    & 45.9    & 6.7    & 29.1    & 30.6    & 47.5 & 10.9 \\  \hline

Source \cite{wu2019Resnet38} & \multirow{3}{*}{\rotHalf{ResNet-38}}      & - & 70.0 & 23.7 & 67.8 & 15.4 & 18.1 & 40.2          & 41.9  & 25.3 & 78.8 & 11.7  & 31.4 & {\ul 62.9} & 29.8 & 60.1 & 21.5 & 26.8 & 7.7 
& 28.1  & 12.0 & 35.4 & - \\

CBST \cite{zou2018unsupervised}        &  & $S_T$      & 86.8            & 46.7 & 76.9            & 26.3          & 24.8          & 42.0    & { \ul 46.0} & \textbf{38.6}          & 80.7          & 15.7          & 48.0          & 57.3          & 27.9          & 78.2          & 24.5            & \textbf{49.6}  & 17.7          & 25.5          & 45.1            & 45.2    & 9.8      \\ 
CRST \cite{zou2019crst}        &  & $S_T$      & 84.5            & 47.7 & 74.1            & 27.9          & 22.1          & \textbf{43.8}    & \textbf{46.5} & {\ul 37.8}          & 83.7          & 22.7          & 56.1          & 56.8          & 26.8          & 81.7          & 22.5            & 46.2 & \textbf{27.5}          & \textbf{32.3}          & {\ul 47.9}            & 46.8 & 11.4          \\ \hline


Source \cite{chen2018deeplab} & \multirow{5}{*}{\rotHalf{DeepLab-v2}}      & -& 71.7      & 	18.5      & 	67.9      &	17.4	      &10.2	      &36.5	      &27.6	      &6.3      &	78.4	      &21.8	      &67.6	      &58.3	      &20.7 	      &59.2       &	16.4       &	12.5      &	7.9       &	21.2       &	13.0 & 33.8 & -\\
MRENT \cite{zou2019crst}        &  &  $S_T$      &  91.8   & 53.4   & 80.6   & \textbf{32.6}   & 20.8   & 34.3   & 29.7   & 21.0   & 84.0   & 34.1   & {\ul 80.6}   & 53.9   & 24.6   & 82.8   & 30.8   & 34.9   & 16.6   & 26.4   & 42.6   & 46.1 & 12.3\\
Ours (DRSL)     &  & $A_I + S_T$    & \textbf{92.8}  &  \textbf{57.5}  & 82.8  & 28.7  & 17.7  & 40.6  & 34.3  & 27.0  & 85.5  & \textbf{42.7}  & 77.8  & 62.3  &  30.8  & 82.2  & 24.3  & 38.5  &  8.4  & {\ul 31.1}  & 39.6  & {\ul 47.6} & {\ul 13.8}\\
Ours (DRSL+)     &  & $A_I + S_T$    & {\ul 92.6}    & {\ul 55.9}    & 82.4    & 29.0    & 24.6    & {\ul 42.7}    & 38.3    & 35.7    & 85.5    & 39.5    & 77.0    & \textbf{64.2}    & 26.2    & 83.9    & 19.5    & 31.6    &  9.3    & 27.1    & 42.5    & \textbf{47.8} & \textbf{14.0}
\\ 

\hline

\end{tabular}
}
\label{table:gta2city}
\end{table*}
\begin{figure*}[t]
	\centering
	\includegraphics[width= \textwidth]{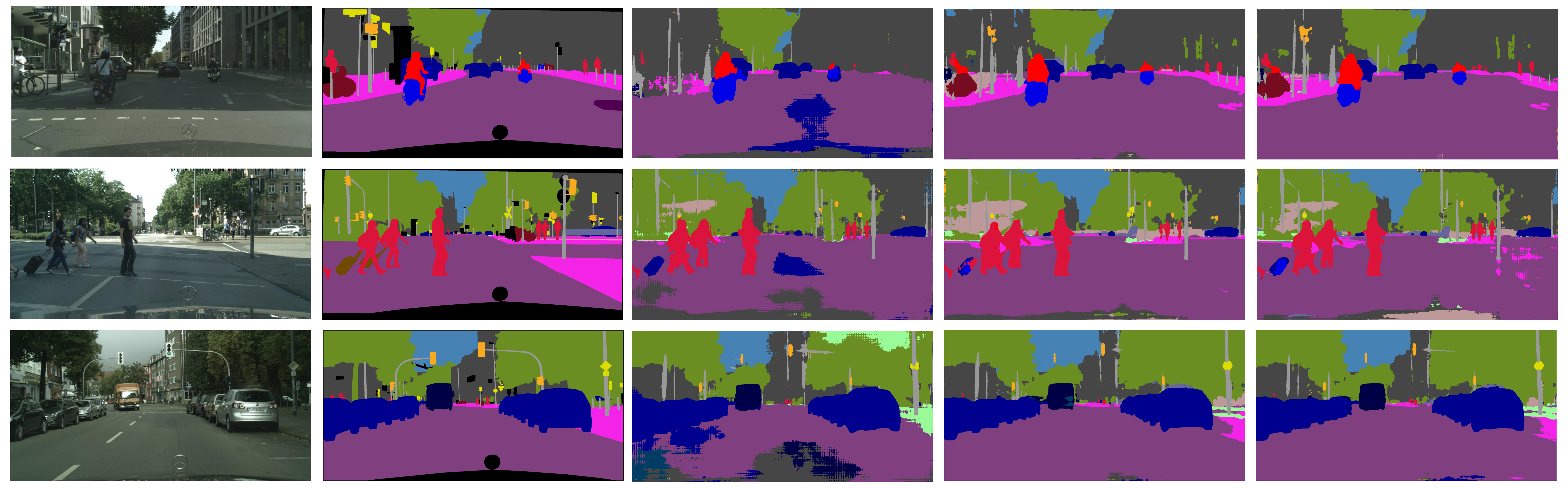}\\
	\footnotesize
	\begin{tabular}{P{2cm}P{2cm}P{2cm}P{2cm}P{2cm}}
    Target Image & Gound Truth & DeepLab-v2 \cite{chen2018deeplab} &DRSL (Ours) & DRSL+ (Ours)
    \end{tabular}
    \caption{Semantic segmentation qualitative results for Cityscapes validation set when adapted from GTA-V dataset.}
	\label{img:gta2city}
\end{figure*}
\section{Experiments and Results}
We performed multiple experiments for domain adaptation of semantic segmentation and compare the obtained results with state-of-the-art methods. 
\subsection{Experimental Setup}
\textbf{Datasets: }
Following \cite{Lian_2019_pycda, zou2018unsupervised, mlsl2020}, we use the standard benchmark setting of \textit{synthetic-to-real} setup for our experiments. Specifically we setup for, \textit{GTA-V to Cityscapes} and \textit{SYNTHIA to Cityscapes} dataset, where the prior is source domain dataset and the later is the target domain dataset. 

\noindent \textbf{Cityscapes} \cite{Cordts2016Cityscapes} dataset is a known benchmark for the task of semantic segmentation and domain adaptation. 
The dataset have 5000 high resolution labeled images partitioned as, training (2975), validation (500) and testing (1125). However, the annotations are only available for training and validation sets.
%
\noindent \textbf{GTA-V} dataset \cite{Richter_2016_ECCV} is obtained from the video game and the images are densely labeled with similar classes to cityscapes. There are 24966 images with spatial resolution spatial resolution $1052 \times 1914$. The GTA-V dataset also covers the road scene imagery. 
\noindent \textbf{SYNTHIA} \cite{Ros_2016_CVPR} is another synthetic labeled images collection having 16 classes similar to Cityscapes. The dataset have 9400 images each with a spatial size $760 \times 1280$. Contrary to GTA-V and Cityscapes, SYNTHIA dataset has more viewpoint variations where; the camera is not supposed to be on the top of a vehicle every time.

\textbf{Network Architecture: }
\label{sec:network}
Following \cite{vu2019advent, tsai2018learning}, we use ResNet-101 \cite{he2016deep} backbone based DeepLab-v2 \cite{chen2018deeplab} as our baseline segmentation model. 
Parallel to the segmentation head is the multi-modal distribution learning based feature regularization (MMDL-FR) module consisting of a combination of DML based Embedding Block (EB) and multi-modal distribution learning. 
We call the DeepLab-v2 last block as the encoder (base network) and the output feature-map as base features. 
For segmentation, these features are passed to segmentation layer while for MMDL-FR, these features are passed to the embedding block(Fig. \ref{img:model}). 
%
%
The embedding block consists of 4 fully convolutional layers with different dilation rates (similar to ones used in the segmentation layer of the segmentation network), producing an aggregated output. 
Unlike \cite{repmet2019}'s fully-connected layers based DML for embedding generation, our strategy preserves the spatial structure necessary for segmentation and requires much less memory. 
The modes of the multi-modal are modeled with a fully connected layer as described in Sec. 3.2. and shown in Fig. \ref{img:model}.
%
For each input the embedding block of the MMDL-FR module outputs an embedding volume $E$ of size $(h\times w\times \hat{d})$. 
For an input image, we select a maximum of $T_e$ embedding vectors per-class at random for further processing.

\textbf{Implementation Details: }
\label{sec:implement-details}
To implement the proposed approach and conduct the experiments, we use PyTorch deep learning framework and a single GTX 1080ti GPU with a single Core-i5 machine with 32GB RAM. The ImageNet \cite{russakovsky2015imagenet} trained weights for ResNet-101 \cite{he2016deep} are used to train the DeepLab-v2 on source dataset. SGD optimizer with weight decay of $5\times 10^{-4}$, momentum of 0.9, and initial learning rate of $2.5 \times 10^{-4}$ for source domain training and $5 \times 10^{-5}$ during adaptation is used. In both source training and adaptation, we used a scale variance (0.5-1.5) and horizontal flipping randomly. For DML and mixture models based classification, the loss weights are set to $\beta = 0.25$ and $\eta = 0.1$ to limit the excessive gradient flow to segmentation model. Similarly for mixture models, the number of modes $M$ is set to 3, and the number of embedding $T_e$ per-class per-image is set to 300. For both source and target domain images, due to GPU memory limitations, small patches of size $512 \times 512$ cropped at random compared to original high-resolution images are processed.  

The baseline segmentation model and the MMDL-FR module are initially trained with original source domain images, \textcolor{black}{in-general called as source-only model.} 
For self-supervised domain adaptation, selection of pixels as pseudo-labels is an important step as the adaptation process depends on the quality of pseudo-labels. We adapt an approach similar to \cite{zou2018unsupervised}, to generate pseudo-labels using the original source data trained model. For a given class $c$, we select $\delta$ confident pixels as pseudo-labels in the first round ($\delta=20\%$ )  and increase this number of pixels ratio by 5\% in each additional round.
To further help the adaptation, we have obtained the translated version of the source domain datasets using CycleGan\cite{hoffman2017cycada} and use these alongside original source images during adaptation. 

\begin{table*}[h]
\centering
\caption{Semantic segmentation performance of DRSL for SYNTHIA to Cityscapes adaptation. 
We present the mIoU (16-classes) and mIoU* (13-classes) comparison with existing state-of-the-art domain adaptation methods for the Cityscapes validation set. 
}

\resizebox{\textwidth}{!}{
\begin{tabular}{l|c|c|cccccccccccccccc|c|c}
\hline 
\multicolumn{19}{c}{SYNTHIA $\rightarrow$ Cityscapes}\\
\hline
Methods     & \rot{Baseline} & \rot{Appr.} & \rot{Road}  & \rot{Sidewalk} & \rot{Building} & \rot{Wall}  & \rot{Fence} & \rot{Pole}  & \rot{T. Light} & \rot{T. Sign} & \rot{Veg.} &  \rot{Sky}   & \rot{Person} & \rot{Rider} & \rot{Car}    & \rot{Bus}    & \rot{M.cycle} & \rot{Bicycle} & \rot{mIoU} & \rot{mIoU*}  \\ \hline \hline
Source \cite{chen2018deeplab} & \multirow{8}{*}{\rotHalf{DeepLab-v2}}  & -        & 64.3  & 21.3  & 73.1  & 2.4  & 1.1  & 31.4  & 7.0  & 27.7  & 63.1  & 67.6  & 42.2  & 19.9  & 73.1  & 15.3  & 10.5  & 38.9  & 34.9  & 40.3  \\
CLAN  \cite{clan_2019_CVPR}        & & $A_O$      & 81.3 & 37.0     & 80.1     & -    & -     & -    & 16.1          & 13.7         & 78.2       & 81.5 & 53.4   & 21.2  & 73.0 & 32.9 & {\ul 22.6}       & 30.7    & -    & 47.8  \\
Structure \cite{structure_2019_CVPR} & & $A_F + A_O$      & \textbf{91.7} & \textbf{53.5}     & 77.1     & 2.5  & 0.2   & 27.1 & 6.2           & 7.6          & 78.4       & 81.2 & 55.8   & 19.2  & 82.3 & 30.3 & 17.1       & 34.3    & 41.5 & 48.7  \\
LSE  \cite{LSE_2020_Naseer}     & & $S_T$       & {\ul 82.9}    & {\ul 43.1}    & 78.1    & 9.3    & 0.6    & 28.2    & 9.1     & 14.4    & 77.0    & 83.5    & 58.1    & 25.9    & 71.9    & \textbf{38.0}    & \textbf{29.4}    & 31.2    & 42.6    & 49.4\\
CRST  \cite{zou2019crst}     & & $S_T$       & 67.7    &  32.2    &  73.9    &  10.7    &  {\ul 1.6}    &  37.4    &  22.2    &  31.2    &  80.8    &  80.5    &  60.8    &  {\ul 29.1}    &  {\ul 82.8}    &  25.0    &  19.4    &  45.3    &  43.8    &  50.1 \\ \hline
Source \cite{wu2019Resnet38} & \multirow{3}{*}{\rotHalf{ResNet-38}}  & -        & 32.6 & 21.5     & 46.5     & 4.81 & 0.03  & 26.5 & 14.8          & 13.1         & 70.8       & 60.3 & 56.6   & 3.5   & 74.1 & 20.4 & 8.9        & 13.1    & 29.2 & 33.6  \\
CBST  \cite{zou2018unsupervised}     & & $S_T$       & 53.6 & 23.7     & 75.0     & 12.5 & 0.3   & 36.4 & {\ul 23.5}          & 26.3         & 84.8       & 74.7 & \textbf{67.2}   & 17.5  & \textbf{84.5} & 28.4 & 15.2       & \textbf{55.8}    & 42.5 & 48.4  \\
MLSL \cite{mlsl2020}   & & $S_T$       &73.7	&34.4	&78.7	&{\ul 13.7}	&\textbf{2.9}	&36.6	&\textbf{28.2}	&22.3	&\textbf{86.1}	&76.8	&{\ul 65.3}	&20.5	&81.7	&31.4	&13.9	&47.3	&44.4	&50.8 \\
\hline
Source \cite{chen2018deeplab} & \multirow{3}{*}{\rotHalf{DeepLab-v2}} & -       & 69.2   & 26.6   & 66.5   &  6.5   &  0.1   & 33.2   &  4.1   & 18.0   & 80.5   & 80.0   & 55.3   & 15.1   & 67.5   & 20.1   &  6.8   & 14.0   & 35.2   & 40.3\\
DRSL   & & $A_I + S_T$       & 70.1     & 30.1     & \textbf{81.6}     & \textbf{15.6}     &  1.0     & {\ul 40.9}     & 20.9     & \textbf{36.4}     & {\ul 85.4}     & {\ul 84.0}     & 59.4     & 26.9     & 81.8     & {\ul 35.9}     & 16.7     & 48.1      &{\ul 45.9}     & {\ul 52.0}\\

DRSL+   & & $A_I + S_T$       & 82.8      & 40.1      & {\ul 81.3}      & 13.0      &  1.6      & \textbf{41.6}      & 19.8      & {\ul 33.1}      & 85.3      & \textbf{84.3}      & 59.5      & \textbf{30.1}      &
 78.6      & 25.3      & 19.8      & {\ul 51.7}      & \textbf{46.7}       &\textbf{53.2} \\ \hline
\end{tabular}
}
\label{table:syn2city}
\end{table*}

\begin{figure*}[t]
	\centering
	\includegraphics[width= \textwidth]{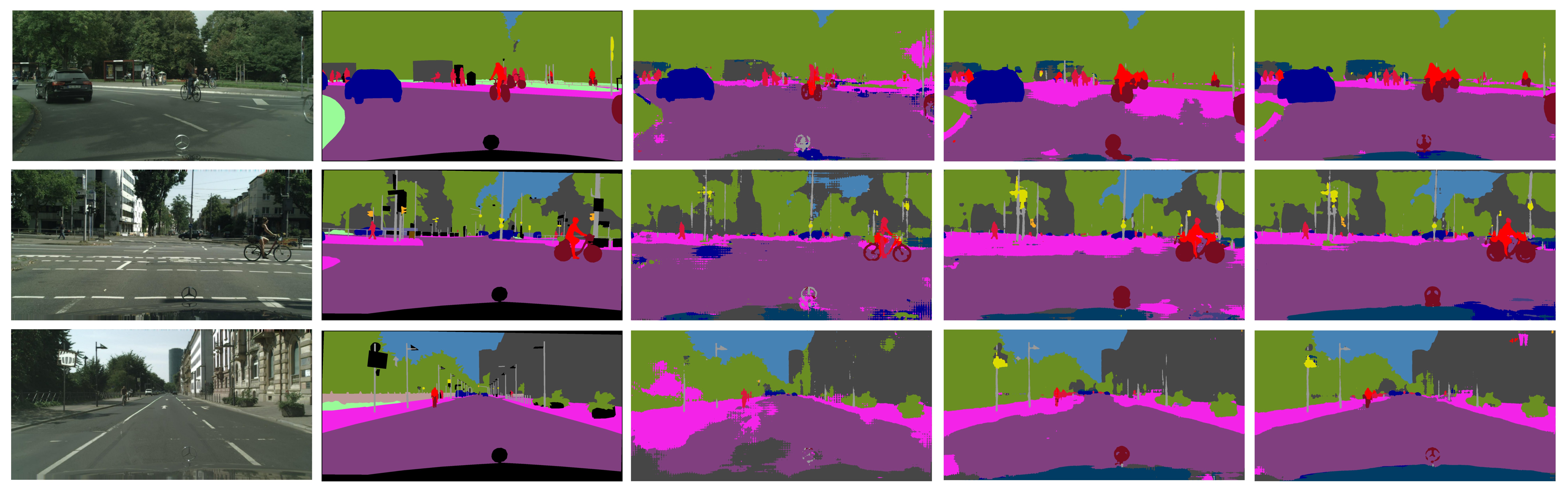}\\
	\footnotesize
	\begin{tabular}{P{2cm}P{2cm}P{2cm}P{2cm}P{2cm}}
    Target Image & Gound Truth & DeepLab-v2 \cite{chen2018deeplab} &DRSL (Ours) & DRSL+ (Ours)
    \end{tabular}
    \caption{Semantic segmentation qualitative results for SYNTHIA to Cityscapes adaptation.}
	\label{img:syn2city}
\end{figure*}

\subsection{Experimental Results}
In this section, we present experimental results of the proposed approach for semantic segmentation. We follow the standard synthetic to real adaptation setup. 

\subsubsection{Results on GTA-V to Cityscapes Adaptation}
Table \ref{table:gta2city} presents domain adaptation performance for the task of semantic segmentation of the proposed DRSL approach compared to existing adversarial learning and self-supervised learning architectures. To have a fair comparison, the methods are divided into three groups where each comparing model is listed with its respective source model and backbone network.
Fig. \ref{img:gta2city} shows example images to highlight the performance of the proposed DRSL qualitatively. The DRSL improves the performance for both objects and stuff classes, as shown in Fig. \ref{img:gta2city} (Column. 4). Small and far away objects like person, traffic light, and signboards are better adapted alongside near to camera objects and large area stuff classes like road, bus, and sidewalk. 
The cross domain mode alignment loss further penalizes the adaptation for small objects, further improving the performance for classes like bicycle, traffic sign, traffic light, pole, fence and person as shown in Table. \ref{table:gta2city} (DRSL+).

Overall, the proposed DRSL+ outperforms the latest self-supervised learning frameworks with clear gaps, surpassing the source dataset trained model with 14.0$\%$ gain in mIoU(last column of Table. \ref{table:gta2city}). 
The DRSL+ performs well on both object classes as well as stuff classes compared to previous methods which may perform better on some classes but fail on other classes.  
Compared to CRST and MRENT \cite{zou2019crst} which regularizes the labels and models for high predictions, the proposed approach achieves a mIoU gain of 1.0 and 1.7\% respectively. Similarly, the DRSL outperforms the PyCDA \cite{Lian_2019_pycda}, which works on pyramid level labeling, and LSE \cite{LSE_2020_Naseer} which incorporates scale invariances with class balancing strategies augmented with higher mIoU baseline models. Compared to composite adversarial learning-based methods like FCAN \cite{zhang2018fcan} and IntraDA \cite{pan2020unsupervised}, DRSL shows improvement with a minimum of 1\% in mIoU and specifically with high margins in small objects. 
Similarly, compared to CAG-UDA\cite{zhang2019category} (mIoU=43.9\% without warm-up training), the DRSL+ gains 3.9\% in mIoU. 

\subsubsection{Results on SYNTHIA to Cityscapes Adaptation}
Table \ref{table:syn2city} presents the proposed DRSL approach segmentation performance for SYNTHIA to Cityscapes adaptation. To have a fair comparison with existing methods, the comparing methods are divided into three groups and the respective source model results with different setups are shown. Moreover, for SYNTHIA to Cityscapes, we show the mIoU (16-classes) and mIoU* (13-classes) as shown by \cite{mlsl2020, zou2018unsupervised}. 
Fig.\ref{img:syn2city} shows qualitative results for DRSL and DRSL+ compared to baseline results. Row-1 and row-2 of Fig.\ref{img:syn2city} focuses on objects like rider, bicycle, person, and the stuff classes,  row-3 highlights the faraway objects and segmentation for road scene imagery.

The DRSL approach performs well on both stuff and object classes adaptation and shows an improvement of 11.7\% in mIoU and 12.9\% in mIoU* compared to the baseline model (source). Compared to strong CBST\cite{zou2018unsupervised} and MLSL\cite{mlsl2020} self-supervised learning approaches, the DRSL shows a minimum improvement of 2.3\% and 2.4\% in mIoU and mIoU* respectively. Similarly, the DRSL shows significant improvement to existing regularization based models, like CRST \cite{zou2019crst} and entropy-based methods, e.g., LSE\cite{LSE_2020_Naseer} and MinEnt \cite{vu2019advent}. Compared to CAG-UDA\cite{zhang2019category} (44.5\% mIoU and 51.4\% mIoU*), the DRSL+ gains 2.2\% in mIoU and 1.9\% in mIoU* respectively. The gaps can be more visible if compared with "without warm-up" training CAG-UDA.   


\subsubsection{Ablation Experiments}
Ablation experiments are performed for GTA-V to Cityscapes.

\label{sec:ablation}
\noindent \textbf{Multi-Modal Distribution Learning based Regularization Module (MMDL-FR): }
During training and adaptation it's essential to understand the balance between the segmentation and different elements of MMDL-FR. 
We search over a range of values to identify (empirically) optimal values for the loss scaling factors, $\beta$ and $\eta$ (Table. \ref{table:cfr-values}).
Based on the experiments, $\beta$ and $\eta$ are set to 0.25 and 0.1 respectively, for all the experiments including SYNTHIA to Cityscapes.
\begin{table}[!htb]
\small
\centering
\caption{Effect of $(\beta, \eta)$ values of the MMDL-FR module.}
\resizebox{3.5in}{!}{
\begin{tabular}{c|ccccc}
\hline
$\beta, \eta$& (0.0, 0.0) & (0.1, 0.1)  &(0.25, 0.1)  & (0.5, 0.5) & (1.0, 1.0)\\ \hline
DRSL (mIoU)    & 44.9 & 46.1    &\textbf{47.6} &45.9 & 46.0\\ \hline
\end{tabular}
}
\label{table:cfr-values}
\end{table}

\noindent \textbf{Effect of MMDL-FR Module on Adaptation Process: }
As described in Sec. \ref{sec:drsl} and Fig. \ref{img:model}, the MMDL-FR module regularizes the encoder (base-network) of the segmentation model with DML based embedding block and MMDL based classification. The MMDL-FR overall enhances the adaptation performance compared to the non-regularized version of the proposed method as shown in Table. \ref{table:drsl-effect}.

\begin{table}[!htb]
\small
\centering
\caption{Effect of MMDL-FR module on adaptation.}
\resizebox{3.5in}{!}{
\begin{tabular}{c|ccc}
\hline
Methods & Source~\cite{chen2018deeplab}  &Without MMDL-FR  & With MMDL-FR \\ \hline
mIoU    & 33.6   &44.9     & \textbf{47.6} \\ \hline
\end{tabular}
}
\label{table:drsl-effect}
\end{table}


\noindent \textbf{Effect of Modes: }
As described in Sec. \ref{sec:drsl}, it is very critical to select correct number of modes for multi-modal in MMDL. We have experimented with multiple number of modes (Table. \ref{table:drsl-modes}) and selected M=3 for all the experiments.

\begin{table}[!htb]
\small
\centering
\caption{Effect to number of modes (M) in MMDL.}
\begin{tabular}{c|ccc}
\hline
Number of Modes (M) & M=1  &M=3  & M=5 \\ \hline
mIoU    & 44.7   &\textbf{47.6}     & 46.2 \\ \hline
\end{tabular}
\label{table:drsl-modes}
\end{table}


\noindent \textbf{Effect of Labels Reduction for MMDL-FR Module: }
The output of the embedding block in the MMDL-FR module is 8 times reduced compared to input image size. 
Embeddings needed to be upsampled 8 times if labels are not reduced requiring a lot of memory. Contrary to this, reducing labels 8 times introduces boxing effect. Based on these observations the scale factor 2 is used. A comparative performance of labels reduction is shown in Table. \ref{table:embedding}.
\begin{table}[H]
\footnotesize
\centering
\caption{Effect of label reduction ratio on mIoU.}
\begin{tabular}{ccccc}
\hline
\multicolumn{5}{c}{GTA-V $\rightarrow$ Cityscapes} \\
\hline
Label Reduction Ratio & 1 & 2 & 4  & 8 \\
Embeddings Upsampling Ratio & 8 & 4 & 2  & 1 \\
Adaptation Performance (mIoU) & 47.1  & \textbf{47.6}   & 46.8    & 46.4   \\
\hline
\end{tabular}
\label{table:embedding}
\end{table}

\noindent \textbf{Pseudo-label Accuracy: }
To understand how the MMDL-FR  results in more accurate pseudo-labels during the adaptation process, we compute mIoU of pseudo-labels for when MMDL-FR is not used (A) and when MMDL-FR is used (B). 
At the start of adaptation (round-0), we have same mIoU for both A $\&$ B (Table-\ref{table:pl-ious}) since MMDL-FR will start to contribute when adaptation starts , i.e., \textit{during} round-0.
Due to MMDL-FR, the predictions by B after round-0 have much lower self-entropy and pseudo-labels have higher mIoU than the ones generated by model-A, thus improving self-supervised domain adaptation. 
\begin{table}[!htb]
\small
\centering
\caption{Pseudo-labels with $\&$ without  MMDL-FR module} 
\resizebox{3.5in}{!}{
\begin{tabular}{c|c|c|c|c}
\hline
\multirow{2}{*}{Method}& \multicolumn{2}{c}{Start of Round-0} & \multicolumn{2}{|c}{Start of Round-1}\\ \cline{2-5}
& mIoU & Self-Entropy & mIoU & Self-Entropy \\ \hline
A: Without MMDL-FR \{ST, ISA\}    & 73.9 &6.56 $\times 10^{-2}$ & 76.4 &1.57$\times 10^{-2}$\\ 
B: With MMDL-FR \{ST, ISA, MMDL-FR\}    & 73.9& 6.56$\times 10^{-2}$& \textbf{78.7}& \textbf{1.14}$\boldsymbol{\times 10^{-2}}$\\ \hline
\end{tabular}
}
\label{table:pl-ious}
\end{table}


\noindent \textbf{Effect of Consistency Loss Weight: }
The cross domain mode consistency loss helps to make the embeddings of the source and target images belonging to the same mode of the same class closer, helping to better adapt the small object classes. However, its contribution in the whole loss needs to be limited to make the system stable. Our experiments suggests $\gamma=0.1$ suits the DRSL+ as shown in Table. \ref{table:drsl-cons-loss}.

\begin{table}[!htb]
\small
\centering
\caption{Effect of cross domain mode consistency loss.}
\begin{tabular}{c|ccc}
\hline
Loss weight $\gamma$ & 0.01  &0.1  & 0.25 \\ \hline
mIoU    & 46.0   &\textbf{47.8}     & 45.3 \\ \hline
\end{tabular}
\label{table:drsl-cons-loss}
\end{table}


\noindent \textbf{Effect of Input Space Adaptation (ISA): }
Removing ISA module, mIoU decreases 1.6 points, from 47.6 (DRSL) to 46.0 (DRSL w/o ISA), indicating that ISA is needed but not vital for the effectiveness of the proposed model.


\section{Conclusion}
\label{sec:conclusion}
In this paper, we propose a distribution regularized self-supervised learning approach for domain adaptation of semantic segmentation. 
Parallel to the semantic segmentation decoding head, we employ a clustering based feature regularization (MMDL-FR) module.
Where segmentation head identifies what can differentiate a class, MMDL-FR explicitly  models intra-class pixel-level feature variations, allowing the model to capture much richer representation of the class at pixel-level, thus improving model's generalization.
Moreover, this disentanglement of information w.r.t tasks improves task dependent representation learning and allows performing separate domain alignments. 
Shared base-network enables MMDL-FR to act as regularizer over segmentation head, thus reducing the noisy pseudo-labels. Extensive experiments on the standard synthetic to real adaptation show that the proposed DRSL outperforms the state-of-the-art approaches. 

{\small
\bibliography{egbib}
}

\end{document}